\documentclass[runningheads]{llncs}
\usepackage{graphicx}
\usepackage{wrapfig}
\usepackage{tikz}
\usepackage{bbding}
\usepackage{pifont}
\usepackage{comment}
\usepackage{amsmath,amssymb} % define this before the line numbering.
\usepackage{color}
\usepackage{subfiles}
\usepackage[pagebackref,breaklinks,colorlinks]{hyperref}
\hypersetup{citecolor=[RGB]{119,185,0}}
\usepackage{xcolor}
\usepackage{colortbl}
\usepackage{hhline}
\usepackage{multirow}
\usepackage{makecell}
\usepackage[width=122mm,left=12mm,paperwidth=146mm,height=193mm,top=12mm,paperheight=217mm]{geometry}
\usepackage{booktabs}
\usepackage{amssymb}
\usepackage{longtable}
\usepackage[misc]{ifsym}

\begin{document}

\pagestyle{headings}
\mainmatter

\title{Cross-Enhancement Transformer for Action Segmentation}

\author{
    Jiahui Wang\textsuperscript{\rm 1},
    Zhengyou Wang\textsuperscript{\rm 1,2 \Letter},
    Shanna Zhuang\textsuperscript{\rm 1,2},
    Hui Wang\textsuperscript{\rm 2}
}
\institute{
    $^1$Shijiazhuang Tiedao University \\
    $^2$Hebei Key Laboratory for Electromagnetic Environmental Effects and Information Processing \\
}

%\end{comment}
%******************
\newcommand{\fn}[1]{\footnotesize{#1}}
\newcommand{\green}[1]{\textcolor[RGB]{96,177,87}{#1}}
\newcommand{\gbf}[1]{\green{\bf{\fn{(#1)}}}}
\def \LN {\eta\xspace}
\def\ie{\textit{i.e.}}
\def\eg{\textit{e.g.}}
\def\etc{etc}
\def\etal{\textit{et al.}}

\newcommand{\wwh}[1]{\textcolor{blue}{[wwh: #1]}}

\authorrunning{J. Wang et al.}

\begin{comment}
\titlerunning{Vision Transformer Adapter for Dense Predictions}
\authorrunning{Zhe Chen et al.}
\institute{Princeton University, Princeton NJ 08544, USA \and
Springer Heidelberg, Tiergartenstr. 17, 69121 Heidelberg, Germany
\email{lncs@springer.com}\\
\url{http://www.springer.com/gp/computer-science/lncs} \and
ABC Institute, Rupert-Karls-University Heidelberg, Heidelberg, Germany\\
\email{\{abc,lncs\}@uni-heidelberg.de}}
\end{comment}
%******************
\maketitle

\begin{abstract}
	Temporal convolutions have been the paradigm of choice in action segmentation, which enhances long-term receptive fields by increasing convolution layers. However, high layers cause the loss of local information necessary for frame recognition. To solve the above problem, a novel encoder-decoder structure is proposed in this paper, called Cross-Enhancement Transformer. Our approach can be effective learning of temporal structure representation with interactive self-attention mechanism. Concatenated each layer convolutional feature maps in encoder with a set of features in decoder produced via self-attention. Therefore, local and global information are used in a series of frame actions simultaneously. In addition, a new loss function is proposed to enhance the training process that penalizes over-segmentation errors. Experiments show that our framework performs state-of-the-art on three challenging datasets: 50Salads, Georgia Tech Egocentric Activities and the Breakfast dataset.
\end{abstract}


%\item 
%\item 
%\item 

% Keywords
\begin{keywords}
	Action segmentation~~ Self-attention mechanism~~ Temporal structure~~ Transformer
\end{keywords}

% Main text
\section{Introduction}\label{Indro}

\par{
	Video action segmentation and classification for untrimmed videos of complex activities which requires to label each frame in a long video by an action class. It has been a hot topic in human action analysis, which is widely used in video surveillance \cite{collins2000introduction}, action teaching, and robotics \cite{vo2014stochastic}.
	Recently, some works \cite{lea2017temporal,farha2019ms,fayyaz2020sct,wang2020gated,li2021efficient} have studied the long range dependencies between correlated actions in action segmentation using temporal convolution networks (TCNs) for models. The TCNs enhance long-term receptive fields by increasing convolution layers. However, as the depth of the convolutional layers increases, the fine-grained information required for frame recognition will be missing. 
}
\par{
	The novel transformer architecture \cite{vaswani2017attention} has led to a big leap forward in capabilities for sequence-to-sequence modeling in NLP tasks. Transformer is famous for using self-attention to extract long-term dependencies in data features. The great transformation of Transformer in NLP has attracted special attention of computer vision. The great transformation of Transformer in NLP has attracted special attention in computer vision, hoping to use Transformer to optimize convolutional neural network-based architectures (CNN) in computer vision tasks. Over the past year, Transformers have enjoyed tremendous success in many computer vision applications, especially in image classification \cite{chen2021crossvit,dosovitskiy2020image,wang2021pyramid,liu2021swin,touvron2021training,liu2022transformer}, video recognition \cite{arnab2021vivit,li2021trear}, video recognition \cite{strudel2021segmenter,zheng2021rethinking}, semantic segmentation \cite{strudel2021segmenter,zheng2021rethinking}, object detection \cite{yi2021asformer}. The Transformer for Action Segmentation (ASFormer) is the first to adopt the transformer architecture in the action segmentation task. ASFormer the explicitly introduced local connectivity inductive and pre-defined hierarchical representation pattern. However, The ASFomer mainly focus on improving hierarchical receptive fields for modeling long-term dependency which is hard-to-excavate the contextual relations between adjacent actions.	
} 
\par{
	The main problem of this work is how to adaptively learn representations from input features to effectively capture global dependencies and the contextual information of adjacent frames. In this paper, we consider using self-attention to enhance the ability of convolution to extract features. Concatenated each layer convolutional feature maps in encoder with a set of features in decoder produced via self-attention, local fine-grained and global information are used in a series of frame actions simultaneously.
}
\par{
	The loss for action segmentation are all trained with frame-level losses, however, these do not adequately penalize sequence-level missclassification. At present, the circle loss provides a paired similarity optimization view of deep feature learning, aiming at maximizing the similarity within the class and minimizing the similarity between classes. We propose to address this over-segmentation by reshaping the Circle Loss \cite{sun2020circle} such that it down-weights the loss assigned to well-classified examples. Our ensemble loss is not only more accurate, but also has a smoothing effect and yields more accurately calibrated sequences. In conclusion, the main contributions of this work are as follows:
}
\par{1. We propose a novel encoder-decoder structure for action segmentation, called Cross-Enhancement Transformer (CETNet) . Our approach can be effective learning of temporal structure representation with interactive self-attention mechanism. Concatenated each layer convolutional feature maps in encoder with a set of features in decoder produced via self-attention, so that it simultaneously exploits both local and global information from a series of frame actions.}
\par{
2. We propose a loss function to enhance the training process and punish over-segmentation. Learning deep features by weighting each similarity score, the loss function has flexible optimization and explicit convergence. Such a loss is highly advantageous in mitigating the effects of over-segmentation and preventing fragmented sequence segmentation. Combining the loss function with a class weighted classification loss function, F1 score can be increased by 5.1\% and segmental edit distance can be increased by 2.3\%.
}
\par{
3. Our approach performs state-of-the-art on three challenging datasets: 50Salads \cite{stein2013combining}, GTEA \cite{fathi2011learning}, and Breakfast \cite{kuehne2014language}. Up to 7.8\% segment F1 score improvement, 3.7\% segment editing distance improvement and 1.9\% accuracy improvement.
}
% Use the style of numbering in square brackets.
% If nothing is used, default style will be taken.
%\begin{enumerate}[a)]
%\item 
%\item 
%\item 
%\end{enumerate}  

%\begin{itemize}
%\item 
%\item 
%\item 
%\end{itemize}  

%\begin{description}
%\item[]
%\item[] 
%\item[] 
%\end{description}  

% Figure
%\begin{figure}[<options>]
%	\centering
%		\includegraphics[<options>]{}
%	  \caption{}\label{fig1}
%\end{figure}
\begin{figure*}
	\centering
	\includegraphics[scale=0.25]{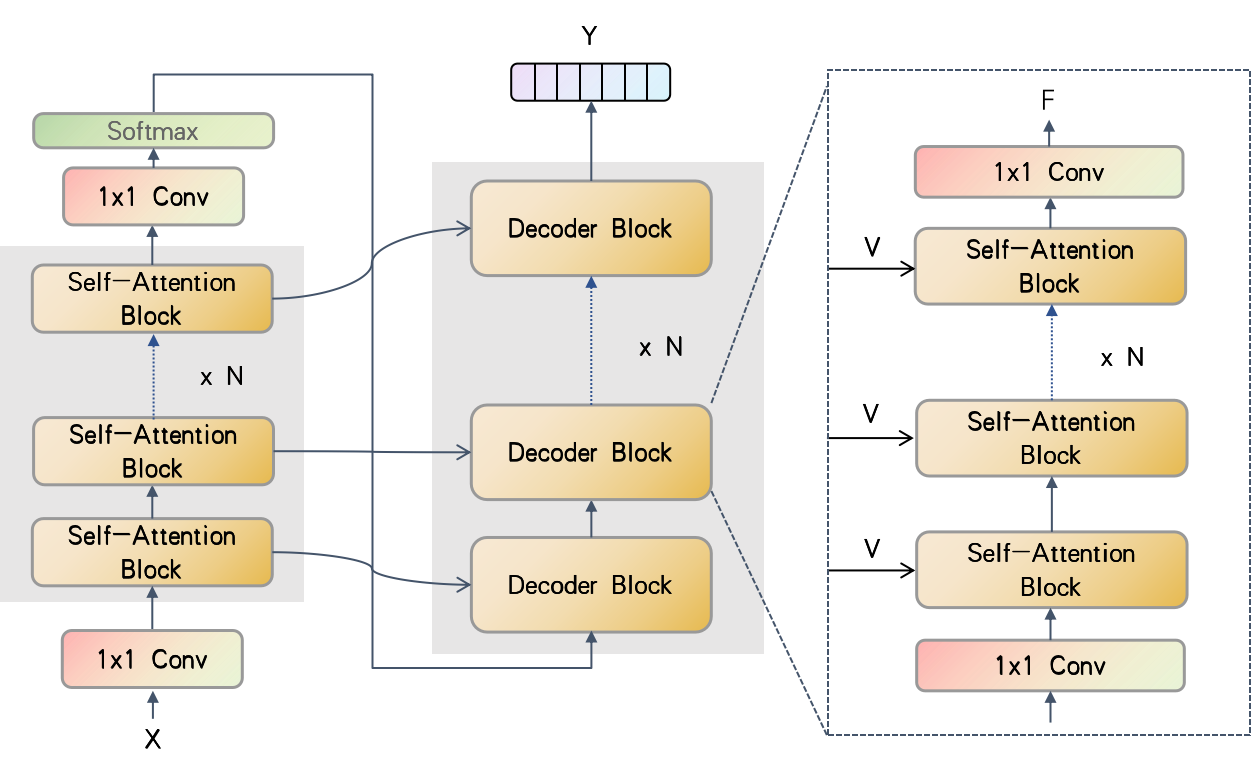}
	\caption{Overview of the Cross-Enhancement Transformer Network (CETNet) comprised of an encoder-decoder architecture with self-attention, which simultaneously exploits both local and global information from a series of frame actions. }
	\label{FIG1}
\end{figure*}
\section{Related work}
\subsection{3D CNN}
\par{
	The 3D CNN-based framework has spatiotemporal modeling capabilities and improves the performance of video action recognition models. 3D ConvNets \cite{tran2015learning,carreira2017quo,feichtenhofer2019slowfast,ming20213d} extended 2D image models \cite{karpathy2014large,wang2018appearance} to the spatial-temporal domain, treating spatial and temporal dimensions in the same way. C3D \cite{tran2015learning} stacked spatiotemporal convolution kernels to efficiently represent video dense structure. I3D \cite{carreira2017quo} extended the convolution and pooling kernels in a very deep image classification network from 2D to 3D to seamlessly learn spatiotemporal features. Our work focuses on frame-level action classification, and video feature extraction is beyond the scope of our work. Following \cite{farha2019ms}, we use I3D \cite{carreira2017quo} for feature extraction as the input to the network, since the videos used for action segmentation are generally long videos that are hard to conduct direct analysis based on raw data.
}
\subsection{Action segmentation}
\par{
	The traditional sliding-window paradigm \cite{karaman2014fast,rohrbach2012database}, which is applied to length and context information, has a long and rich history, and other methods use Markov \cite{kuehne2016end} models or RNN \cite{yue2015beyond} models to apply rough time modeling. Recently, inspired by the success of temporal convolution in speech synthesis \cite{van2016wavenet} , temporal convolutional networks (TCNs) transformed a commonly used architecture for temporal video segmentation. Some TCNS works \cite{lea2017temporal,lei2018temporal,farha2019ms} mainly focus on improving receptive fields that model long-term dependencies with encoder structures, dilated convolutions, or deformable convolutions. \cite{ishikawa2021alleviating,li2021efficient} build architecture on the two-branch approach : One branch exploits wide long-term time receptive fields based on TCNs. The second exploits frame-boundary based on action boundary regression. \cite{yi2021asformer} explore the Transformer on action segmentation task, which introduced inductive local connectivity and a preset hierarchical representation model. The above methods capture long-term dependencies by increasing the depth layers of temporal convolution, which lead to fine-grained novel loss between adjacent frames. An innovative compared to previous methods, our approach uses self-attention mechanism to augment convolutional operators by concatenating the maps of the convolution features of each layer in the encoder with a set of features in the self-attention decoder. And it solves the problem of fine-grained loss in adjacent frames.
}
\section{Method}
\par{
	In the section, we present our CETNet structure for action segmentation. Our CETNet uses an encoder-decoder architecture with self-attention, which simultaneously exploits both local and global information from a series of frame actions, as shown Fig. \ref{FIG1}. The encoder will first capture global temporal information by expanding the layers of the self-attention blocks which use a deep series of dilated convolutions. Then the decoders will use the initialized predictions and hierarchical features, obtained from the encoder, to perform incremental refinement. Finally, the result will be passed to the combined loss function to optimize the frame-level classification. Section \ref{section3.1} illustrates the details of self-Attention block with expanded dilated convolutions. Section \ref{section3.2} shows how to utilize the encoder to capture hierarchical features and a long-term feature extractor. Section \ref{section3.3} introduces our refinement scheme in decoder.Section \ref{section3.4} introduces the combined loss function and training details of our framework.
}
\subsection{Self-Attention block}\label{section3.1}
\par
\begin{figure}
	\centering
	\includegraphics[scale=0.5]{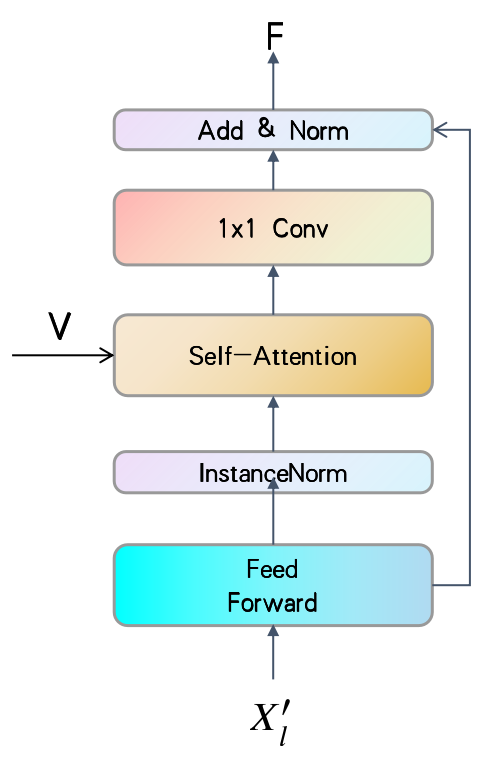}
	\caption{Illustrates the details of self-Attention block with expanded dilated convolutions.}
	\label{FIG2}
\end{figure}

{
	As shown in Fig.~\ref{FIG2}, Given the input features $ X_l'\in {R^{T \times D}}'$ which is extracted from input videos or previous ${l_{th}}$ layer, where ${\rm{D'}}$ is the dimension and $T$ is the video length. The first of the self-attention Block is a feed-forward layer which consists of $1{\rm{D}}$ dilated temporal convolution and ${\rm{RELU}}$ activation. We increase the dilation rates with kernel size $3$ for conducting different temporal receptive fields:
}
\begin{equation}
	{\rm{S}} = \{ {2^i}, i = 0,1,2....\} 
\end{equation}
\noindent where ${\rm{i}}$ is index of Self-Attention block. The receptive field grows exponentially with the number of layers, which helps prevent the model from overfitting the training data \cite{lea2017temporal}. We use instance normalization after feed-forward to improve performance:
\begin{equation}
	X_l{''} = In(FFN({X_l}'))
\end{equation}
\noindent where $ In $ is instance normalzation, $ FFN $ is the feed-forward layer, ${\rm{X'}}$ is input features and ${\rm{X''}}$ is the output after instance normalization.
\begin{equation}
	Q = X_l'{W_q},K = X_l'{W_k},V = X_l'{W_v}
\end{equation}
\begin{equation}
	\begin{aligned}
		Att &= Attention(Q,K,V) \\
		&= soft\max (Q{K^T}/\sqrt {{d_k}})~V
	\end{aligned}
\end{equation}
\noindent where $ {W_q},{W_k},{W_v} \in {R^{T \times (C/r)}} $ are the query, key and value matrices of learnable parameters, $\sqrt {{{\rm{d}}_{\rm{k}}}} $ is the scaling factor, $ C $ and $ r $ are the dimension and hyperparameter. Note that the input of $ V $ is different between encoder and decoder. In the encoder(described in Section \ref{section3.2}) , the input of $ V $ is the same as $ Q $. In the decoders(described in Section \ref{section3.3}), in order to alleviate the information leakage by temporal correlation, we propose to use each encoder hierarchy feature as the $ V $ value to perform incremental optimization. We also include a $1 \times 1$ conv and residual connection after the self-attention operation, as this helps to adjust the dimension for subsequent operations:
\begin{equation}
	{\rm{F}} = LN(\smallint (Att)) + X_l'
\end{equation}
\noindent where $ X_l' $ is the input of encoder or decoder , $ LN $ is Layer normalzationand $\smallint$ is $1 \times 1$ convolution operation.
\subsection{Encoder}\label{section3.2}
\par{
	The encoder consists of N sequential Self-Attention Block layers, We set N=10 in our paper(ablation experiment show in section \ref{section4.5.1}). Before the first layer of encoder, we use a FC layer to reduce the input feature dimension from ${\rm{D}}$ to ${\rm{D'}}$.
}
\begin{equation}
	\begin{aligned}	
		X_{l + 1}' = \phi (X_l')
	\end{aligned}
\end{equation}
\noindent where $\phi$ is a Self-Attention Block discussed in Section \ref{section3.1}. $X_{l + 1}'$ is the next layer of encoder. The last layer of the encoder uses the softmax output as the initial embedding for each frame prediction, which contains an abstract representation of the global features. 
\begin{equation}
	{Y_{\rm{c}}} = soft\max (W{{\rm{X}}_{\rm{l}}} + b)
\end{equation}
\noindent where ${Y_c} \in {R^c}$, $c$ is classify frame-level action classes, $W$ and $b$ are the weights and bias.
\subsection{Decoders}\label{section3.3}
\par{
	Decoder consists of N sequential decoder block that is similar to encoder structure. A little different is that the query $Q$ and key $K$ are obtained by concatenating the encoder output and the preceding layer, while the value $V$ is only obtained from the self-attention of the corresponding layer in the encoder. Inspired by \cite{yi2021asformer}, The self-attention mechanism utilizes training to focus attention weights at each location and continuously refines all locations. Furthermore, since the output of each decoder is an initial prediction with different hierarchy of temporal relationships, the decoder is aligned with the encoder's self-attention layer to continuously optimize the global and local information, reduce fine-grained information loss to prevent over-segmentation errors.
}
\begin{figure*}
	\centering
	\includegraphics[scale=0.25]{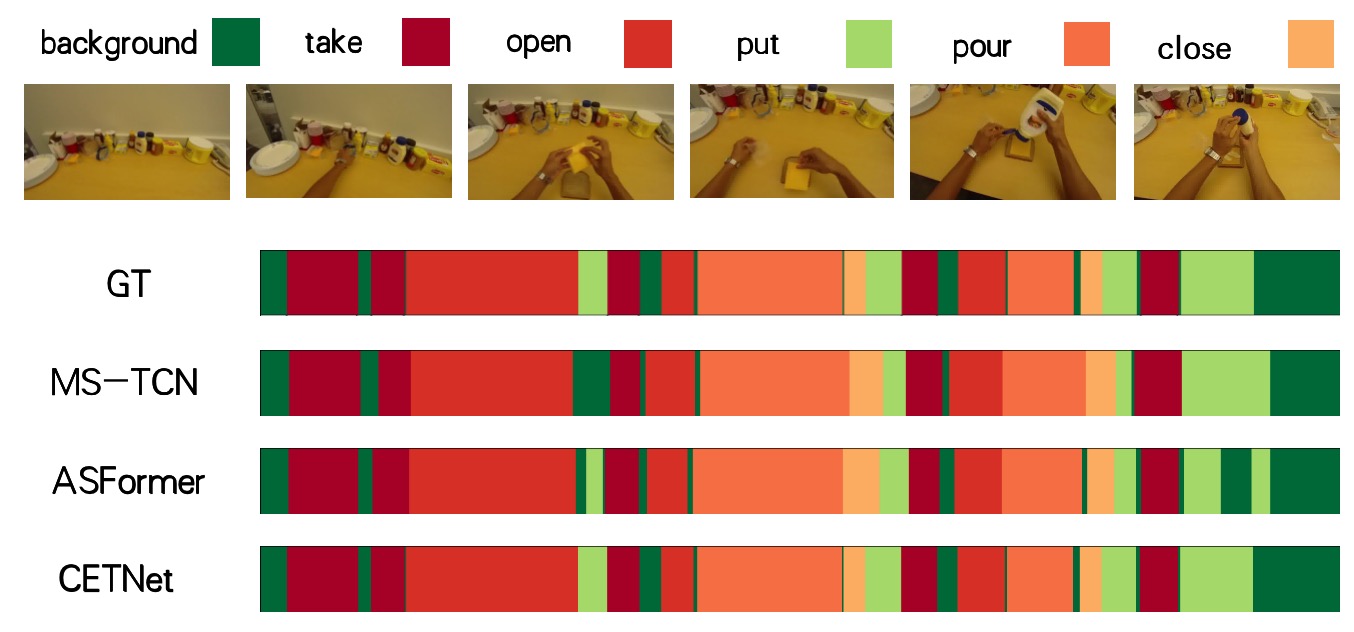}
	\caption{Qualitative result from the GTEA dataset for comparing different method of action segmentation. Only part of the whole video is shown for clarity. We can see that our CETNet method is most closed to groundtruth.}
	\label{FIG3}
\end{figure*}
\subsection{Loss Function and Implementation Details}\label{section3.4}
\subsubsection{Loss Function}
\par{
	Circle loss \cite{sun2020circle} is a loss function that learns deep features by weighting each similarity score, and has flexible optimization and explicit convergence:
}
\begin{equation}
	\begin{aligned}
		{L_{circle}} &= \log [1 + \sum\limits_{i = 1}^K {\sum\limits_{j = 1}^L {\exp (\gamma (\alpha _n^is_n^j - \alpha _p^is_p^j))} } ]\\
		&= \log [1 + \sum\limits_{j = 1}^L {\exp (\gamma \alpha _n^js_n^j} )\sum\limits_{i = 1}^K {\exp ( - \gamma \alpha _p^is_p^i)} ]
	\end{aligned}
\end{equation}
\noindent in which $\alpha _n^i$ and $\alpha _p^i$ are weighting factors, and x is a scale factor. There are similarity scores as $\{ s_p^i\} ~(i = 1,2, \cdot  \cdot  \cdot ,K)$ and $\{ s_n^j\}~ (i = 1,2, \cdot  \cdot  \cdot ,L)$, respectively. $K$ is the similarity scores in the class and $L$ is the similarity score between classes. 
\par{
	When we regard the softmax value in a classification loss function as the probability that the sample belongs to a certain class, constant weight scaling is a common operation. Circle loss has an independent weighting factor, which is multiplied by each similarity score before rescaling. Therefore, optimization is more flexible without the constraint of constant weight scaling. So we use a simple set prediction loss ${L_{loss}}$ to use the given set of frame actions.
}
\begin{equation}
	{L_{loss}} = {{\rm{L}}_{cls}} + \lambda {L_{mse}} + \beta {L_{circle}}
\end{equation}
\noindent where ${{\rm{L}}_{cls}}$ is a cross-entropy loss, ${L_{mse}}$ is the smooth loss in \cite{farha2019ms}. $\lambda $ and $\beta $ are balance weight. All losses of the encoder and decoder are accumulated and trained to search for the minimum optimal value.
\subsubsection{Implementation Details}
\par{
	The final CETNET structure consists of encoder-decoder. The encoder consists of 10 self-attention layers, and the number of decoders corresponds to the number of self-attention layers of the encoder, in other words, the decoder contains 10 decoders, each containing 10 self-attention layers. In all experiments, our deep learning model framework is based on the pytorch framework, and the physical hardware uses two NVIDIA RTX 2080ti GPUs and ubuntu with cuda10.1.	
}
\par{
	In order to prove the effectiveness of our model, we adopt the same preprocessing and super parameters in ASFormer \cite{yi2021asformer}. Keeping the fps of the 50Salads dataset the same as the other datasets, we take a frame step of 2 in 50Salads , and a frame step of 1 in GTEA and Breakfast datasets. We train the model for 120 epochs,batch size is 1 and the kernel size in all layers is 3. For loss hyperparameter setting, we set $\lambda $ = 0.15, $\beta $ = 0.001.
}

\begin{table}[ht]
	\centering
	\caption{Impact of Mulit-head and dimension on the GTEA dataset.}
		\begin{tabular}{ccccccc}
			\toprule
			Multi-head & dim   & \multicolumn{3}{c}{F1@\{10,25,50\}} & Edit  & Acc \\
			\midrule
			1     & 64    & \textbf{91.8} & \textbf{91.2} & \textbf{81.3} & \textbf{87.9} & \textbf{80.3} \\
			2     & 64    & 90.8  & 89.3  & 80.9  & 87.6  & 79.7 \\
			3     & 64    & 90.6  & 88.9  & 79    & 87.1  & 78.8 \\
			4     & 64    & 90.4  & 88.9  & 79.7  & 86.3  & 78.6 \\
			2     & 128   & 90.7  & 89.4  & 80.1  & 87.2  & 79.4 \\
			\bottomrule
	\end{tabular}%
	\label{tab1}%
\end{table}%
\begin{table*}[htbp]
	\centering
	\caption{Impact of the decoders hierarchical refinement on GTEA and 50salads dataset}
	\begin{tabular}{crrrrrrrrrr}
		\toprule
		Dataset & \multicolumn{5}{c}{GTEA}              & \multicolumn{5}{c}{50salads} \\
		\midrule
		Cross-decoder & \multicolumn{3}{c}{F1@\{10,25,50\}} & \multicolumn{1}{l}{Edit} & \multicolumn{1}{l}{Acc} & \multicolumn{3}{c}{F1@\{10,25,50\}} & \multicolumn{1}{l}{Edit} & \multicolumn{1}{l}{Acc} \\
		\midrule
		
		no-cross & 90.4  &  88.3 &  79.5  & 86  &  78.4  & 81  & 79.3    & 71.5  & 73.8  & 83.6 \\				
		ahead-cross  & 77.9 & 74.5  & 67.5 & 69.3  & 79.1    & 54.2  & 53  & 47  & 43.9 & 84.3 \\
		ahead-cross (only) &80.2	&79	&69.3&	73.3&	78.7&	54.3&	52.7&	48.7&	43.1&	85.9  \\
		behind-cross & 90.9  & 89.8  & 80.7  & 86.8  & 79.3  & 86.8    & 85.1  & 79  & 80.9  & 86.3 \\	
		behind-cross (only) & 91.2&	90.4&	81&	86.8&	78.8&	86.1&	85&	77.6&	80.7&	85.7\\
		all-cross  & \textbf{91.8}  & \textbf{91.2} & \textbf{81.3}  & \textbf{87.9} & \textbf{80.3}  & \textbf{87.6} & \textbf{86.5} & \textbf{80.1} & \textbf{81.7} & \textbf{86.9} \\
		\bottomrule
	\end{tabular}%
	\label{tab7}%
\end{table*}%

\begin{table*}[htbp]
	\centering
	\caption{Ablation study of loss function on the GTEA and 50salads dataset.}
	\begin{tabular}{crrrrrrrrrr}
		\toprule
		Dataset & \multicolumn{5}{c}{GTEA}              & \multicolumn{5}{c}{50salads} \\
		\midrule
		Loss & \multicolumn{3}{c}{F1@\{10,25,50\}} & \multicolumn{1}{l}{Edit} & \multicolumn{1}{l}{Acc} & \multicolumn{3}{c}{F1@\{10,25,50\}} & \multicolumn{1}{l}{Edit} & \multicolumn{1}{l}{Acc} \\
		\midrule
		$ {\rm{L}}_{cls} + \lambda {L_{mse}} (\lambda=0.15) $ & 90.3  & 89.4  & 80.5    & 86.2  & 79.6  & 86.3  & 85    & 77.7  & 79.9  & 86 \\
		$ {\rm{L}}_{cls} + \lambda {L_{mse}} (\lambda=0.75) $ & 90.9  & 89.7  & 79.7  & 86.8  & 79  & 85.5    & 83.8  & 76.6  & 79  & 85 \\
		$ {L_{loss}} (\lambda=0.75 , \beta=0.001) $  & 91.3 & 90  & 80.5 & 87.7  & 79.7    & 85.1  & 84.3  & 77.8  & 78.9 & 85.5 \\
		$ {L_{loss}} (\lambda=0.15 , \beta=0.001) $ & \textbf{91.8}  & \textbf{91.2} & \textbf{81.3}  & \textbf{87.9} & \textbf{80.3}  & \textbf{87.6} & \textbf{86.5} & \textbf{80.1} & \textbf{81.7} & \textbf{86.9} \\
		\bottomrule
	\end{tabular}%
	\label{tab4}%
\end{table*}%
\begin{table*}[ht]
	\centering
	\caption{Impact of layers on the GTEA and 50salads dataset.}
	\begin{tabular}{crrrrrrrrrr}
		\toprule
		Dataset & \multicolumn{5}{c}{GTEA}              & \multicolumn{5}{c}{50salads} \\
		\midrule
		Layer (N) & \multicolumn{3}{c}{F1@\{10,25,50\}} & \multicolumn{1}{l}{Edit} & \multicolumn{1}{l}{Acc} & \multicolumn{3}{c}{F1@\{10,25,50\}} & \multicolumn{1}{l}{Edit} & \multicolumn{1}{l}{Acc} \\
		\midrule
		5 & 88.4  & 87.2  & 76    & 83.5  & 76.7  & 57.6  & 54   & 43,7  & 47.4  & 75.4 \\
		6 & 91.2  & 90.1  & 79.1  & 87.4  & 78.3  & 73.1    & 70.3  & 61.2  & 63.8  & 79.7 \\
		7  & \textbf{92.1} & 90.9  & \textbf{81.9} & 87.6  & 79.4    & 79.6  & 77.5  & 69.7  & 72.2 & 82.1 \\
		8 & 90.3  & 88.8  & 77.4  & 86.4  & 77.6  & 83.4  & 82 & 74.4  & 76.6 & 84 \\
		9 & 90 & 89.1 & 77.8  & 85.5  & 79 & 86.1  & 84.3  & 78.3  & 79.8  & 85.7 \\
		10 & 91.8  & \textbf{91.2} & 81.3  & \textbf{87.9} & \textbf{80.3}  & \textbf{87.6} & \textbf{86.5} & \textbf{80.1} & \textbf{81.7} & \textbf{86.9} \\
		\bottomrule
	\end{tabular}%
	\label{tab2}%
\end{table*}%
\section{Experiments}
\subsection{Dataset}
\par{
	50Salads \cite{stein2013combining} dataset consists of 50 videos belonging to 17 action classes. The average length of each video is 6 minutes and contains 20 actions. GTEA \cite{fathi2011learning} dataset consists of 28 videos belonging to 11 action classes.We use four different training-test splitting strategies to guarantee the validity of the experiment. Breakfast \cite{kuehne2014language} dataset consists of 1712 videos belonging to 48 action classes with 18 different kitchens. It is the largest and most challenging dataset in action segmentation. In addition to using 5-fold cross-validation on the 50salads dataset, we use 4-fold cross-validation for evaluation on the other two datasets and report the average results.
}
\subsection{Evaluation metrics}
\par{
	Accuracy (Acc), edit distance (Edit) and F1 scores (F1@{10,25,50}) are three evaluation metrics commonly used in action segmentation. The frame-wise accuracy is the accuracy of the action prediction per frame in a video. However, it is unable to penalize the over-segmentation errors. Edit and F1 scores are the action segmentation metrics used to evaluate whether it is over-segmented. Edit is a measure representing the similarity between predicted and groudtruth. F1 scores represent the scores at different overlap thresholds, which score at {10\%, 25\%, 50\%}, denote by F1@{10,25,50}.
}
\subsection{Impact of multi-head self-attention}
\par{
	In the transformer \cite{vaswani2017attention}, Multi-Head self-attention is used to divide the model into multiple heads to form multiple subspaces, so that the model can focus on different aspects of information and splice the results of multiple projections. The final result is then obtained by linear transformation that enhance the feature transformation. Here, we explore the effect of different self-attention heads on GTEA. Except the last one (the divide heads in 2-head set 64 dimension), all other setting is the same as the single-head attention. According to Tab. \ref{tab1}, we can find that multi-heads are insensitive to our method. This could be an overfitting problem due to the increased number of parameters.
}
\begin{table*}[htbp]
	\centering
	\caption{Comparig our proposed method with existing methods on GTEA and 50salads dataset}
		\begin{tabular}{crrrrrrrrrr}
			\toprule
			Dataset & \multicolumn{5}{c}{GTEA}              & \multicolumn{5}{c}{50salads} \\
			\midrule
			Method & \multicolumn{3}{c}{F1@\{10,25,50\}} & \multicolumn{1}{l}{Edit} & \multicolumn{1}{l}{Acc} & \multicolumn{3}{c}{F1@\{10,25,50\}} & \multicolumn{1}{l}{Edit} & \multicolumn{1}{l}{Acc} \\
			\midrule
			MS-TCN\cite{farha2019ms} & 87.5  & 85.4  & 74.6  & 81.4  & 79.2  & 76.3  & 74    & 64.5  & 67.9  & 80.7 \\
			MS-TCN++\cite{li2020ms} & 88.8  & 85.7  & 76    & 83.5  & 80.1  & 80.7  & 78.5  & 70.1  & 74.3  & 83.7 \\
			SSTDA\cite{chen2020action} & 90    & 89.1  & 78    & 86.2  & 79.8  & 83    & 81.5  & 73.8  & 75.8  & 83.2 \\
			SSTDA+HASR\cite{ahn2021refining} & 90.9  & 88.6  & 76.4  & 87.5  & 78.7  & 83    & 81.5  & 73.8  & 75.8  & 83.2 \\
			BCN\cite{wang2020boundary}   & 88.5  & 87.1  & 77.3  & 84.4  & 79.8  & 82.3  & 81.3  & 74    & 74.3  & 84.4 \\
			C2F-TCN\cite{singhania2021coarse} & 90.3  & 88.8  & 77.7  & 86.4  & \textbf{80.8} & 84.3  & 81.8  & 72.6  & 76.4  & 84.9 \\
			ETSN\cite{li2021efficient} & 91.1  & 90    & 77.9  & 86.2  & 78.2  & 85.2  & 83.9  & 75.4  & 78.8  & 82 \\
			ASRF\cite{ishikawa2021alleviating} & 89.4  & 87.8  & 79.8  & 83.7  & 77.3  & 84.9  & 83.5  & 77.3  & 79.3  & 84.5 \\
			ASFormer\cite{yi2021asformer} & 90.1  & 88.8  & 79.2  & 84.6  & 79.7  & 85.1  & 83.4  & 76    & 79.6  & 85.6 \\
			CETNet(ours) & \textbf{91.8} & \textbf{91.2} & \textbf{81.3} & \textbf{87.9} & 80.3  & \textbf{87.6} & \textbf{86.5} & \textbf{80.1} & \textbf{81.7} & \textbf{86.9} \\
			\bottomrule
	\end{tabular}%
	\label{table5}%
\end{table*}
\subsection{Effect of the decoders hierarchical refinement}
\par{
	To demonstrate that our decoder exploits multiple levels of temporal relationships for refinement, we perform ablation studies that stack different numbers of decoders. We evaluate the importance of global cross level self-attention by changing the number of decoder layers. We compare all-layer cross model compared to defect-layer cross models with different number of decoders on GTEA and 50Salads datasets. As shown in the Table. \ref{tab7}, we explore the cross-attention effect of the first five layers (ahead-cross) and the last five layers (behind-cross), other layers are replaced by the output of the encoder. We also conduct experiments using only the cross-attention of the front and rear five layers. The all-layer cross model utilizes all temporal-level information for refinement and perform the best. By comparing the segment edit distances and F1 scores of these models, we can see that only the previous hierarchy layers in the encoder produce a lot of over-segmentation errors. Deep layers contain more abstract temporal information and the cross-attention of the back layers is better than the previous layers. On the other hand, the cross-enhancement architecture can reduce over-segmentation errors and improve F1 scores. This improvement is clearly visible when all hierarchies are used, greatly improving the segmentation metrics. In the experiment, our cross self-attention structure is better than the commonly used attention mechanism, which does not consider the relationship between the previous encoder layers and decoder outputs predicted by each layer.
}
\subsection{Ablation study on hyper-parameters}
\subsubsection{Ablations of the number of blocks}\label{section4.5.1}
\par{
	Stacking more self-attention blocks can get a larger receptive field (as introduced in Sec. \ref{section3.1}) , but it will cost more computation and memory. We conduct ablation studies on different numbers of self-attention blocks in the encoder-decoder of GTEA and 50salads datasets, as shown in table. \ref{tab2}. Although the two F1 scores in the GTEA dataset achieve the best performance when the self-attention layer is set to 7, it does not perform well in the 50salads dataset. So we set the self-attention block to 10 by default.
}
\subsubsection{Comparing loss functions for the CETNet}
\par{
	As shown in Table. \ref{table4}, compares different hyper-parameters combination of loss functions. Our proposed loss function improves frame-level accuracy, F1 score and edit distance, and achieves the best performance when hyperparameters $\lambda $ = 0.15, $\beta $ = 0.001. Learning deep features by weighting each similarity score, the loss function has flexible optimization and explicit convergence. Such a loss is highly advantageous in mitigating the effects of over-segmentation and preventing fragmented sequence segmentation.
}
\begin{table}[ht]
	\centering
	\caption{Comparing our proposed method with existing methods on Breakfast dataset}
	\setlength{\tabcolsep}{2.5mm}
	\begin{tabular}{cccccc}
		\toprule
		Dataset & \multicolumn{5}{c}{Breakfast} \\
		\midrule
		Method & \multicolumn{3}{c}{F1\{10,25,50\}} & Edit  & Acc \\
		\midrule
		MS-TCN\cite{farha2019ms} & 52.6  & 48.1  & 37.9  & 61.7  & 66.3 \\
		MS-TCN++\cite{li2020ms} & 64.1  & 58.6  & 45.9  & 65.6  & 67.6 \\
		BCN\cite{wang2020boundary}   & 68.7  & 65.5  & 55    & 66.2  & 70.4 \\
		ETSN\cite{li2021efficient}  & 74    & 69    & 56.2  & 70.3  & 67.8 \\
		ASRF\cite{ishikawa2021alleviating}  & 74.3  & 68.9  & 56.1  & 72.4  & 67.6 \\
		SSTDA\cite{chen2020action} & 75    & 69.1  & 55.2  & 73.7  & 70.2 \\
		C2F-TCN\cite{singhania2021coarse} & 76.3  & 69.9  & 54.6  & 74.5  & 70.8 \\
		ASFormer\cite{yi2021asformer} & 76    & 70.6  & 57.4  & 75    & 73.5 \\
		CETNet(ours) & \textbf{79.3} & \textbf{74.3} & \textbf{61.9} & \textbf{77.8} & \textbf{74.9} \\
		\bottomrule
	\end{tabular}%
	\label{table4}%
\end{table}

\subsection{Comparison with the state of the art}
\par{
	In the experiment, we show that our framework performs state-of-the-art on three challenging datasets: 50Salads, GTEA,and Breakfast datasets. As shown in Table. \ref{table5} and Table. \ref{table4}. Our model achieves the state-of-the-art methods on the 50Salads and Breakfast datasets compared to previous work. Our CETNet is having up to 5.4\% and 6.8\% improvement for the segmental F1 score on 50Salads and Breakfast respective. Although the accuracy of the C2F-TCN Method is higher than CETNet on GTEA, the F1 score perform a large margin up to 4.63\% for the F1 score. We visualized the prediction of labels as shown in Fig. \ref{FIG3}.
}
\section{Conclusion}
\par{
	In this paper, we present CETNet, a novel encoder-decoder interactive self-attention mechanism for learning global features, to improve the classification accuracy for action segmentation. To address the effects of over-segmentation and prevent fragmented sequence segmentation, we further develop a loss function to re-weighting each similarity score under supervision. With extensive experiments, we demonstrate that our proposed CETNet outperforms the state-of-the-art models by a large margin on 50Salads, GTEA and Breakfast. While our current work has only scratched the surface of cross-ehancement transformers for action segmentation, we anticipate that more work will be done in the future to develop effective cross-ehancement transformers for other action applications, including action recognition, action assessment, and action correction.
}
% Uncomment and use as the case may be
%\begin{theorem} 
%\end{theorem}

% Uncomment and use as the case may be
%\begin{lemma} 
%\end{lemma}



\section*{Acknowledgement}
This work is supported by the Innovation Research  Funds for Shijiazhuang Tiedao University (No. YC2022057), the National Nature Science Foundation of China (No. 61972267), and the Nature Science Foundation of Hebei Province (No. F2019210306).

\clearpage

\bibliographystyle{splncs04}
\bibliography{arxiv}
\end{document}